\documentclass[sigconf]{acmart}
\usepackage{booktabs}
\usepackage{enumerate}
\usepackage{multirow}
\usepackage{utfsym}
\usepackage{cleveref}
\AtBeginDocument{%
  }

\copyrightyear{2025} 
\acmYear{2025} 
\setcopyright{acmlicensed}\acmConference[MM '25]{Proceedings of the 33rd ACM International Conference on Multimedia}{October 27--31, 2025}{Dublin, Ireland}
\acmBooktitle{Proceedings of the 33rd ACM International Conference on Multimedia (MM '25), October 27--31, 2025, Dublin, Ireland}
\acmDOI{10.1145/3746027.3755694}
\acmISBN{979-8-4007-2035-2/2025/10}

\begin{document}

\title{DACA-Net: A Degradation-Aware Conditional Diffusion Network for Underwater Image Enhancement}


\author{Chang Huang}
\affiliation{%
  \institution{The Hong Kong University of Science and Technology (Guangzhou), Southern Marine Science and Engineering Guangdong Laboratory (Guangzhou)}
  \city{Guangzhou}
  \country{China}}
\email{chuang932@connect.hkust-gz.edu.cn}

\author{Jiahang Cao}
\affiliation{%
  \institution{The Hong Kong University of Science and Technology (Guangzhou)}
  \city{Guangzhou}
  \country{China}
}
\email{jcao248@connect.hkust-gz.edu.cn}

\author{Jun Ma}
\affiliation{%
  \institution{The Hong Kong University of Science and Technology (Guangzhou)}
  \city{Guangzhou}
  \country{China}
}
\email{eejma@hkust-gz.edu.cn}

\author{Kieren Yu}
\affiliation{%
  \institution{The Hong Kong University of Science and Technology (Guangzhou)}
  \city{Guangzhou}
  \country{China}
}
\email{kyu219@connect.hkust-gz.edu.cn}

\author{Cong Li}
\affiliation{%
  \institution{The Hong Kong University of Science and Technology (Guangzhou)}
  \city{Guangzhou}
  \country{China}
}
\email{cli166@connect.hkust-gz.edu.cn}

\author{Huayong Yang}
\affiliation{%
  \institution{Southern Marine Science and Engineering Guangdong Laboratory (Guangzhou)}
  \city{Guangzhou}
  \country{China}
}
\email{yanghy@gmlab.ac.cn}

\author{Kaishun Wu}
\authornote{Corresponding author}
\affiliation{%
  \institution{The Hong Kong University of Science and Technology (Guangzhou)}
  \city{Guangzhou}
  \country{China}
}
\email{wuks@hkust-gz.edu.cn}

\settopmatter{authorsperrow=4}






\begin{abstract}
Underwater images typically suffer from severe colour distortions, low visibility, and reduced structural clarity due to complex optical effects such as scattering and absorption, which greatly degrade their visual quality and limit the performance of downstream visual perception tasks. Existing enhancement methods often struggle to adaptively handle diverse degradation conditions and fail to leverage underwater-specific physical priors effectively.
In this paper, we propose a degradation-aware conditional diffusion model to enhance underwater images adaptively and robustly. Given a degraded underwater image as input, we first predict its degradation level using a lightweight dual-stream convolutional network, generating a continuous degradation score as semantic guidance. Based on this score, we introduce a novel conditional diffusion-based restoration network with a Swin UNet backbone, enabling adaptive noise scheduling and hierarchical feature refinement. To incorporate underwater-specific physical priors, we further propose a degradation-guided adaptive feature fusion module and a hybrid loss function that combines perceptual consistency, histogram matching, and feature-level contrast.
Comprehensive experiments on benchmark datasets demonstrate that our method effectively restores underwater images with superior colour fidelity, perceptual quality, and structural details. 
Compared with SOTA approaches, our framework achieves significant improvements in both quantitative metrics and qualitative visual assessments.
\end{abstract}





\maketitle

\section{Introduction}

The ocean is a vital ecosystem essential to human survival, playing a pivotal role in ecosystem monitoring~\cite{obura2019coral, edinger2000reef}, climate regulation~\cite{hughes2017global, hoegh2023ocean}, biodiversity conservation~\cite{tittensor2010global, mallet2014underwater}, and resource exploration~\cite{danovaro2017deep, ferrari2016quantifying}. 
With increasing exploration and utilisation of marine resources, underwater robots, autonomous underwater vehicles, and marine monitoring networks have been extensively deployed in marine ecological assessment, subsea infrastructure inspection, and resource surveys. 
Due to its significant advantages, such as intuitiveness, information richness, ease of interpretation, technological maturity, and low implementation cost, optical-based underwater vision remains the predominant approach for marine environmental monitoring.
In these scenarios, obtaining high-quality underwater imagery is critical for accurate object detection, scene interpretation, and effective decision-making. 
However, the complex optical characteristics of aquatic environments severely degrade imaging quality: First, wavelength-dependent absorption causes red light (620-750 nm) to attenuate over 90\% within just 1 meter depth, while blue-green light (450-570 nm) penetrates farther, leading to severe color distortion and bias~\cite{jobson1997multiscale}. 
Second, scattering effects (Mie and Rayleigh scattering) from suspended particles reduce contrast, obscure details, and introduce haze-like artifacts~\cite{levin2007closed}.
These degradations pose significant challenges to subsequent computer vision algorithms. 
Therefore, developing robust underwater image enhancement (UIE) techniques to generate clear, realistic, and informative images is crucial for advancing intelligent marine perception systems.


Despite significant advances in underwater image enhancement technologies in recent years, several critical challenges persist.
Current underwater image enhancement methods are primarily divided into traditional physical model-based methods and deep learning-based approaches. 
Traditional methods generally rely on optical models to estimate underwater parameters, such as the Dark Channel Prior-based dehazing models~\cite{he2010single} and underwater light attenuation models~\cite{chiang2011underwater}. 
However, such methods depend on idealised assumptions (e.g., uniform illumination, static water) and often fail in real-world scenarios because of their failure in precise parameter estimation. 
Recent advances in deep learning, particularly generative adversarial networks~\cite{goodfellow2014generative} and convolutional neural networks~\cite{li2020underwater}, have improved enhancement performance through data-driven strategies.
For example, supervised learning methods such as WaterNet~\cite{li2019underwater}, WaterGAN~\cite{li2017watergan} and Ucolor~\cite{li2021underwater} achieve enhancement by learning to pair synthetic degraded images with real ones.
Although these methods achieve notable results by training on synthetic degraded-ground truth image pairs, they suffer from limited generalisation due to substantial domain gaps between synthetic and real-world images.
Diffusion models~\cite{ho2020denoising}, as an emerging generative paradigm, have demonstrated potential in image restoration tasks due to their progressive denoising mechanism and high sample diversity.
However, conventional diffusion models employ fixed noise schedules, which cannot adapt to the spatiotemporal variations of degradation levels in underwater environments. 
For instance, foreground regions obstructed by suspended particles require aggressive denoising, while distant areas demand depth-specific colour correction. 
Moreover, existing methods lack an interpretable analysis of the temporal dynamics during denoising, leaving model optimisation without theoretical guidance and hindering the balance between computational efficiency and restoration quality.
Therefore, there is a compelling need to develop more generalised, adaptable underwater image enhancement frameworks and provide deeper insights into the enhancement process, offering substantial theoretical significance and practical utility.

\textbf{Contributions:} In this paper, we propose a degradation-guided conditional diffusion framework for underwater image enhancement, aiming to effectively address the diverse degradations in underwater environments. 
Our approach leverages a reversible denoising process guided by a degradation score, which adaptively adjusts the noise schedule and feature normalisation across the diffusion steps. 
To enhance the perceptual fidelity and semantic consistency of restored images, we incorporate physical imaging priors into the network via a Physical-guided Fusion Module (PGFM), which includes red-channel compensation and frequency-aware attention. 
In addition, we design a hybrid loss function that jointly considers perceptual similarity, colour distribution consistency, and feature alignment through the integration of perceptual loss, histogram loss, and contrastive loss, promoting visually realistic and semantically faithful restoration.
Moreover, we embed a lightweight dual-stream degradation estimation network that predicts image-specific degradation scores in a data-driven yet interpretable manner. 
Guided by both learned degradation priors and underwater physical characteristics, the proposed method improves visual restoration quality and ensures robustness across diverse underwater scenarios. 
The multi-scale Swin-UNet backbones and condition-aware architectures collaborate to accomplish controlled enhancements with high fidelity and semantic integrity.
Overall, our main contributions can be summarised as follows:
\begin{itemize}

    \item  We propose a degradation-aware conditional diffusion restoration framework, which adaptively guides the denoising process using an estimated degradation score, significantly enhancing the adaptability and controllability of restoration under varying underwater conditions.
    
    \item  We design a dual-stream lightweight degradation estimator, which learns to regress PSNR-based degradation scores from paired images for continuous conditioning, effectively providing accurate and interpretable degradation cues for subsequent diffusion enhancement.

    \item  We introduce PGFM and a hybrid loss function that collectively enhance color fidelity, perceptual consistency, and fine-grained detail recovery by incorporating underwater priors and multi-level constraints.

    \item Qualitative and quantitative results demonstrate that the proposed DACA-Net achieves superior performance, reaching the state-of-the-art level on multiple underwater image datasets.

\end{itemize}

\section{Related Work}

\subsection{Underwater Image Enhancement}

Existing underwater image enhancement methods can generally be categorised into traditional physics-based methods and deep learning-based methods. 

\subsubsection{Physics-based Methods.}
He et al.~\cite{he2010single} proposed the Dark Channel Prior, which effectively removes haze and colour distortion by estimating atmospheric light and transmission maps.
Additionally, Li et al.~\cite{li2015underwater} developed a colour correction-based approach to enhance underwater images. 
Besides, Derya and Tali~\cite{akkaynak2019sea} introduced an adaptive sea-bottom image correction technique, significantly improving image contrast in underwater scenes.
Also, Chiang et al.~\cite{chiang2011underwater} integrated underwater light attenuation models with wavelength compensation to recover red-channel information from blue-green channels,
These approaches aimed to improve images by estimating atmospheric light, adjusting colour distribution, and enhancing contrast, but they demonstrated limited performance in handling complex underwater scenarios.
The unique optical properties of seawater, such as differential attenuation of light wavelengths~\cite{akkaynak2018revised, pope1997absorption, jerlov1976marine, kirk1994light}, result in blue-green images with substantial loss of critical detail information, presenting higher requirements for image enhancement technologies.
\subsubsection{Deep Learning-based Methods.}
With the rapid development of deep learning technologies, neural network-based underwater image enhancement methods have demonstrated immense potential. 
For instance, recent studies explored unsupervised approaches to underwater image restoration~\cite{fu2022unsupervised}.
Fabbri et al.~\cite{fabbri2018enhancing} presented an underwater image quality enhancement framework based on GANs, innovatively utilising adversarial training strategies to recover image details.
Li et al.~\cite{li2019underwater} comprehensively studied the application of deep learning in underwater image enhancement, establishing the first deep learning benchmark for underwater image enhancement. 
UAGAN by~\cite{wang2023underwater} significantly improved underwater image visual quality through adversarial training strategies. 
Raaveendran et al.~\cite{raveendran2021underwater} conducted an in-depth analysis of critical challenges in deep learning-based underwater image enhancement, systematically summarising the limitations of existing methods. 
Zhang et al.~\cite{zhang2021underwater} proposed a novel deep generative adversarial network, offering new insights for processing underwater scenes.

Notably, these methods still face critical challenges: 
(1) Underwater scene data acquisition is extremely difficult, with models often relying on limited and potentially non-representative synthetic data;
(2) Existing methods universally suffer from domain gap issues, which struggle to 
 be effectively generalised to real, complex underwater scenarios;
(3) There is a lack of profound theoretical explanations for underwater image degradation mechanisms.

\subsection{Diffusion Models}

Diffusion probabilistic models have recently emerged as a powerful generative paradigm, effectively capturing complex image data distributions by progressively injecting and subsequently removing noise. 
Initially introduced by Ho et al.~\cite{ho2020denoising} and further expanded by Song et al.~\cite{song2020score} through the score-based generative modelling framework, diffusion models have demonstrated notable performance in various vision tasks. 
Recent studies have validated their remarkable ability to restore colour fidelity, enhance perceptual quality, and recover structural details from severely degraded images~\cite{kulikov2023sinddm, sahoo2024diffusion, lugmayr2022repaint}.
Nevertheless, when applied to underwater scenarios, current diffusion-based methods still exhibit critical limitations.
The conventional diffusion frameworks typically employ fixed noise scheduling strategies, lacking the flexibility required to cope with diverse and complex underwater degradation conditions. 
Moreover, existing diffusion methods rarely incorporate explicit physical priors inherent in underwater imaging~\cite{guan2023diffwater, tang2023underwater}, resulting in less robust and interpretable enhancement outcomes. Furthermore, the intrinsic dynamics of the diffusion processes often remain opaque, limiting the interpretability and controllability of the enhancement results. These shortcomings highlight opportunities for further improvement in diffusion-based underwater image enhancement methods.

\section{Methodology}
This section introduces the detailed methodology of the proposed framework. 
~\Cref{3.1} provides an overview and the problem formulation of our approach. 
~\Cref{3.2}  presents the dual-stream degradation estimator used for predicting degradation scores. 
~\Cref{3.3}  describes the conditional diffusion-based restoration process. 
~\Cref{3.4}  elaborates on the physical-guided fusion module, and ~\Cref{3.5}  outlines the hybrid loss functions in this work.

\subsection{Overview}
\label{3.1}
\begin{figure*}[h]

        \centering
        \includegraphics[width=\linewidth]{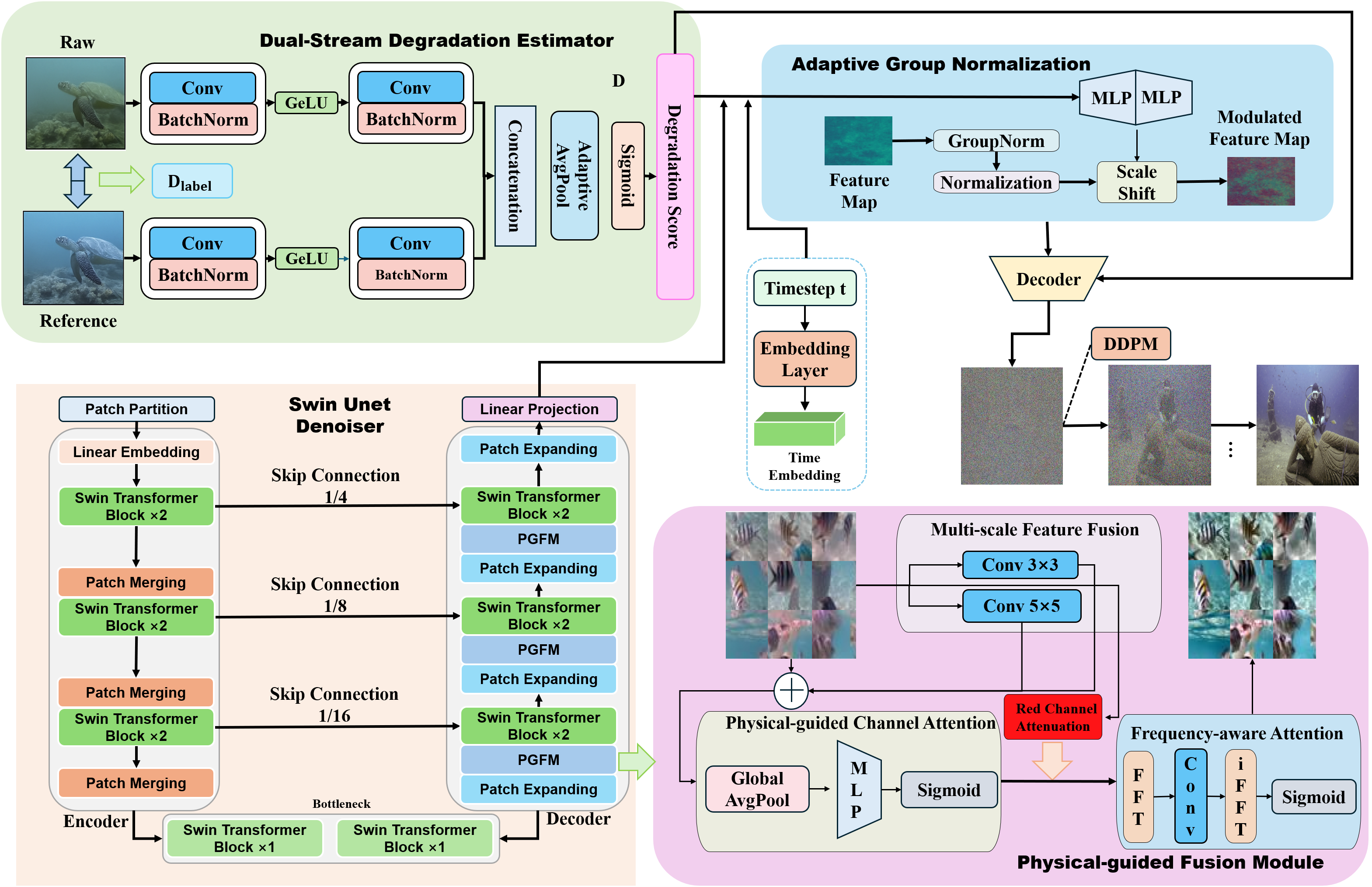 } 
    \caption{Overview illustration of the proposed DACA-Net framework. DACA-Net is composed of four main components: Dual-Stream Degradation Estimator, AdaGN, Swin UNet Denoiser and PGFM. Given a raw image and its reference, the degradation estimator predicts a global degradation score to guide the restoration. This score, together with timestep embeddings, is injected into the Swin UNet-based conditional diffusion network via AdaGN for degradation-aware denoising. The PGFM module is embedded within the denoising network to enhance colour fidelity and detail restoration using physical priors and multi-scale cues. The final enhanced image is generated through the DDPM process.}
    \label{Overview}
\end{figure*}

This section presents an overview of the proposed DACA-Net framework, as illustrated in~\Cref{Overview}. 
The framework consists of four main components: a Dual-Stream Degradation Estimator, an AdaGN module, a Swin-Unet Denoising Backbone, and a PGFM module.
To enable degradation-aware conditional diffusion enhancement, we first design a lightweight dual-stream convolutional network to regress a degradation score $\mathbf{D}$ from paired raw and reference images, supervised by the PSNR. 
This score serves as a global conditioning signal to guide the denoising trajectory.
The denoising process is carried out by a Swin-Unet backbone, which embeds the PGFM module in its decoder pathway. 
During decoding, the network incorporates both the degradation score $\mathbf{D}$ and timestep $\mathbf{t}$ via the AdaGN module, which adaptively modulates the feature normalization and noise scheduling, achieving semantic-aware restoration.
To further boost detail fidelity and perceptual consistency, the PGFM module integrates physical priors and frequency-aware enhancement to recover fine-grained structures. 
The entire process follows a conditional DDPM formulation, reconstructing the enhanced image from noise under the joint guidance of $\mathbf{D}$ and $\mathbf{t}$, demonstrating robustness across diverse underwater degradation scenarios.

\subsection{Dual-Stream Degradation Estimator}
\label{3.2}

To enable controllable enhancement in our diffusion framework, we design a lightweight degradation estimation network that predicts a scalar degradation score, as shown in~\Cref{Overview}. 
This score serves as a global condition throughout the denoising process.
We formulate degradation estimation as a regression task using paired images: the degraded underwater image and its corresponding clean reference. 
To ensure both efficiency and expressiveness, we employ a dual-stream CNN architecture with depthwise separable convolutions~\cite{chollet2017xception}, which significantly reduces computational cost while preserving feature representation capability.
The two input images are processed in parallel by shared-weight convolutional encoders. 
Their extracted features are then concatenated and passed through a shallow regression head to produce a scalar degradation score:
\begin{equation}
    D = \mathcal{R} \left( \phi(\mathbf{I}_{\text{raw}}), \phi(\mathbf{I}_{\text{ref}}) \right)
\end{equation}
where $\mathbf{I_{raw}}$ and $\mathbf{I_{ref}}$denote the degraded underwater image and its corresponding clean reference, respectively; $\mathbf{\phi(\cdot)}$ denotes the depthwise CNN feature extractor and $\mathbf{\mathcal{R}(\cdot)}$ is a fusion and regression module.

We employ Peak Signal-to-Noise Ratio (PSNR) as the reference-based image quality measure to generate ground-truth labels for degradation estimation. 
Given the clean reference and degraded image, the ground-truth degradation score is computed as:
\begin{equation}
    D_{\text{label}} = 1 - \frac{\text{PSNR} - \text{PSNR}_{\min}}{\text{PSNR}_{\max} - \text{PSNR}_{\min}}
\end{equation}
where the PSNR value is normalized to $\mathbf{[0,1]}$ range based on empirical dataset-specific minimum and maximum values.

\subsection{Conditional Diffusion-based Restoration}
\label{3.3}

Diffusion probabilistic models~\cite{ho2020denoising, moser2024diffusion, kulikov2023sinddm, wang2023stylediffusion} have emerged as a powerful class of generative models that generate high-quality data by reversing a gradual noising process. Given a clean image $\mathbf{x_0}$, the forward diffusion process gradually adds Gaussian noise over $\mathbf{T}$ steps, producing a sequence $\left\{ \mathbf{x}_t \right\}_{t=1}^{T}$ according to:
\begin{equation}
    q(\mathbf{x}_t \mid \mathbf{x}_{t-1}) = \mathcal{N} \left( \mathbf{x}_t ; \sqrt{1 - \beta_t} \, \mathbf{x}_{t-1}, \, \beta_t \mathbf{I} \right)
\end{equation}
where $\beta_t \in (0, 1)$ is a predefined noise variance schedule. After sufficient steps, the image becomes nearly pure Gaussian noise $\mathbf{x}_T \sim \mathcal{N}(0, \mathbf{I})$.

The reverse process, parameterized by a neural network $\mathbf{\epsilon_\theta}$ aims to denoise $\mathbf{x_T}$ step-by-step to recover $\mathbf{x_0}$. 
The reverse denoising step is formulated as:
\begin{equation}
    p_{\theta}(\mathbf{x}_{t-1} \mid \mathbf{x}_t) = \mathcal{N} \left( \mathbf{x}_{t-1} ; \mu_{\theta}(\mathbf{x}_t, t), \sigma_t^2 \mathbf{I} \right)
\end{equation}
where the mean $\mu_{\theta}(\mathbf{x}_t, t)$ is predicted using the estimated noise $\epsilon_{\theta}(\mathbf{x}_t, t)$ and the known diffusion parameters. The restoration output $\hat{\mathbf{x}}_0$ is obtained by denoising from $\mathbf{x_T}$ down to $\mathbf{x_0}$ in $\mathbf{T}$ steps.
In our framework, the input is a raw underwater image, which is embedded as $\mathbf{x_T}$ and denoised under the guidance of the predicted degradation score using the conditional network. 
The final output $\hat{\mathbf{x}}_0$ is the enhanced image.

To achieve controllable and degradation-aware enhancement for underwater images, we propose a conditional diffusion-based restoration framework. 
The model is built upon a Swin-UNet~\cite{liu2021swin, cao2022swin} backbone, which efficiently captures multi-scale hierarchical features.
More importantly, we introduce a degradation score $D \in [0, 1]$ as a global conditioning signal, enabling both noise modulation and feature adaptation throughout the denoising process:
\begin{equation}
    \tilde{\beta}_t = \beta_t \cdot (1 + \alpha D)
\end{equation}
where $\mathbf{\beta_t}$ is the original noise parameter at diffusion step $\mathbf{t}$, $\mathbf{\alpha}$ is a learnable scaling coefficient. 
This formulation allows the model to adaptively control the noise injection based on the estimated degradation level, offering semantic-aware denoising capacity. 
In addition, we integrate the degradation score into the feature transformation pipeline via a lightweight Adaptive Group Normalisation (AdaGN) module, which extends GroupNorm by learning scale and shift parameters from the conditioning signal~\cite{wu2018group} as illustrated in~\Cref{Overview}.
Given an input feature map $\mathbf{\mathbf{x} \in \mathbb{R}^{N \times C \times H \times W}}$, AdaGN operates as:
\begin{equation}
    \text{AdaGN}(\mathbf{x}, D) = \gamma(D) \cdot \frac{\mathbf{x} - \mu_G(\mathbf{x})}{\sigma_G(\mathbf{x}) + \epsilon} + \beta(D)
\end{equation}
where $\mathbf{\mu_G(x)}$ and $\mathbf{\sigma_G(x)}$ are the mean and standard deviation computed over each group of channels; $\mathbf{\gamma(D)}$ and $\mathbf{\beta(D)}$ are scale and shift parameters predicted from the degradation score using a two-layer MLP.
AdaGN enables fine-grained control of feature distribution across spatial and channel dimensions by modulating the feature statistics using degradation-aware parameters. 
This mechanism improves both the visual fidelity of restoration and the robustness to varying underwater degradation levels.

\subsection{Physical-guided Fusion Module}
\label{3.4}

To effectively integrate physical priors into the enhancement process, we introduce a Physical-guided Fusion Module (PGFM) as illustrated in~\Cref{Overview}. 
This module is designed to enhance colour fidelity and fine detail restoration by leveraging two key strategies: red channel compensation and frequency-aware attention.

Underwater images often suffer from severe red-channel attenuation due to wavelength-dependent light absorption. 
To address this, we introduce a learnable red-channel amplification mechanism within the intermediate feature space.
Given an input feature map $\mathbf{{F \in \mathbb{R}^{ N \times C \times H \times W}}}$, we scale the red channel (indexed by 0) using a degradation-aware modulation:
\begin{equation}
    \mathbf{F}_{\text{red}} = \mathbf{F}[:, 0, :, :] \cdot (1 + \gamma \cdot D)
\end{equation}
where $\mathbf{D}$ is the predicted degradation score and $\gamma$ is a learnable coefficient. 
The amplified red channel is then concatenated with the original feature map along the channel dimension to form a physically informed representation. 
This operation encourages the network to restore colour consistency lost due to red light absorption.

To complement colour correction, PGFM further enhances fine structural details via a frequency-aware attention mechanism~\cite{liu2023fair}. Specifically, given an intermediate feature map $\mathbf{F}$, we apply a Fast Fourier Transform to convert it to the frequency domain, where an attention map is learned and applied before reconstructing the enhanced spatial features.
The full process can be formulated as:
\begin{equation}
\mathbf{F}_{\text{freq}} = \text{IFFT} \left( \mathcal{F}(\mathbf{F}) \cdot \sigma \left( \text{Conv}(\mathcal{F}(\mathbf{F})) \right) \right)
\end{equation}
where $\mathbf{\mathcal{F}(\cdot)}$ and $\mathbf{IFFT(\cdot)}$ denote the FFT and inverse FFT operations, respectively; $\mathbf{Conv(\cdot)}$ is a convolutional layer for learning frequency-domain weights; and 
$\mathbf{\sigma(\cdot)}$is the sigmoid function.

\subsection{Loss Function}
\label{3.5}

The design of loss functions is critical for Underwater Image Enhancement.
Traditional pixel-wise losses such as L1 or L2 losses often fail to effectively capture perceptual details and global stylistic features~\cite{zhao2016loss}. 
We propose a multi-component composite loss function that comprehensively considers perceptual similarity, histogram matching, and feature contrast.
Specifically, given the predicted enhanced image $\mathbf{I_{pred}}$ and its corresponding reference image $\mathbf{I_{ref}}$, we define the loss terms as follows.

To correct colour shifts in underwater images, we introduce a channel-wise histogram alignment loss based on KL divergence. For each RGB channel, we discretise pixel value distributions of the predicted and target images into 32 bins and compute their KL divergence:
\begin{equation}
    \mathcal{L}_{\text{hist}} = \sum_{c \in \{R, G, B\}} \mathrm{KL}\left( H_c(I_{\text{pred}}) \,\|\, H_c(I_{\text{ref}}) \right)
\end{equation}
where $\mathbf{H_c(\cdot)}$ denotes the normalized histogram probability. 
Softmax normalisation and a smoothing term are applied for numerical stability. 
This loss is inspired by colour transfer methods~\cite{reinhard2001color, afifi2019else} but incorporates dynamic bin adaptation to handle non-uniform underwater colour attenuation.

Inspired by the groundbreaking work~\cite{gatys2015neural}, we utilise a pre-trained VGG16~\cite{simonyan2014very} to extract deep semantic features, assessing image perceptual similarity by comparing differences in feature space. 
\begin{equation}
    \mathcal{L}_{perc} = \|\phi_{\text{VGG}}(I_{pred}) - \phi_{\text{VGG}}(I_{ref})\|_{1}
\end{equation}
where $\mathbf{\phi_{VGG}}$ represents the feature extractor of the first 16 layers. 
By matching deep features~\cite{johnson2016perceptual}, this loss effectively preserves textures and suppresses artifacts.

Besides, motivated by contrastive learning frameworks~\cite{chen2020simple, sohn2016improved}, this loss maximises cosine similarity between enhanced and target images in feature space:
\begin{equation}
    \mathcal{L}_{contra} = 1 - \frac{\phi_{\text{VGG}}(I_{pred}) \cdot \phi_{\text{VGG}}(I_{ref})}{\|\phi_{\text{VGG}}(I_{pred})\| \cdot \|\phi_{\text{VGG}}(I_{ref})\|}
\end{equation}
By enhancing semantic consistency at the feature level, this loss mitigates mode collapse caused by degradation-level variations~\cite{brock2018large}.
The total loss is a weighted sum of the components:
\begin{equation}
    \mathcal{L}_{total} = \lambda_1\mathcal{L}_{perc} + \lambda_2\mathcal{L}_{hist} + \lambda_3\mathcal{L}_{contra}
\end{equation}

\section{Experiments}
This section presents the experimental settings and results to validate the effectiveness of the proposed method. 
\Cref{4.1} provides implementation details, including datasets, evaluation metrics, and training configuration. 
\Cref{4.2} introduces the baseline methods used for comparison. 
\Cref{4.3} reports quantitative and qualitative results across various benchmarks.
Finally, \Cref{4.4} conducts extensive ablation studies to analyse the contribution of each key component in our framework.

\begin{table*}[t]
\caption{Quantitative comparison of degradation estimation regression networks in terms of accuracy and efficiency.}
\centering
\begin{tabular}{c|ccc|ccc}
\hline
    Method & MSE $\downarrow$ & MAE $\downarrow$ & Pearson Coefficient(\%)  $\uparrow$ & Param.(G) & GFLOPs & Time (ms)\\
\hline
     ResNet34~\cite{he2016deep}  & 0.0029 & 0.0432 & 91.65 & 21.29 & 4.80 & 2.62  \\ 
     AlexNet~\cite{krizhevsky2012imagenet}  & 0.0035 & 0.0476 & 92.17 & 57.01 & 0.92 & 1.91  \\
    GoogLeNet~\cite{szegedy2015going}  & 0.0056 & 0.0614 & 85.79 & 5.60 & 1.97 & 3.59  \\
    ShuffleNet~\cite{zhang2018shufflenet}  & 0.0060 & 0.0626 & 81.04  & 1.25 & \textbf{0.20} & 3.10 \\
    MobileNet~\cite{howard2017mobilenets}  & 0.0057 & 0.0596 & 89.64 &  2.23 & 0.43 & 2.39 \\
    VGG16~\cite{simonyan2014very}  & 0.0029 & 0.0418 & 93.01 & 134.26 & 20.16 & 2.47   \\
    \textbf{Ours}    & \textbf{0.0012} & \textbf{0.0271} & \textbf{96.03}   & \textbf{1.02} & 0.27 & \textbf{1.75}  \\
\hline
\end{tabular}
\label{DegradationEstimator}
\end{table*}

\subsection{Implementation Details}
\label{4.1}
\textbf{Settings.} We implement the whole experiment with PyTorch using a commodity workstation with a single GTX-4080 GPU. 
In our experiments, we set the maximum number of training epochs to 200 and the batch size to 32.
We use the Adam optimiser and set the initial learning rate to 1e-4.
All the images are resized to $\mathbf{256 \times 256}$.
The time step of the diffusion model is set to 1500.

\noindent\textbf{Datasets.}  We conduct experiments on two public underwater image enhancement datasets: UIEB~\cite{li2019underwater} and LSUI~\cite{peng2023u}. 
UIEB contains 890 paired images, where the reference images are obtained through human selection among results from multiple enhancement algorithms. 
We use its standard test split with 90 images for evaluation. 
LSUI is a large-scale dataset that includes 4279 paired underwater images across diverse scenes. 
Following the official setting, we use 3879 image pairs for training and the remaining 400 pairs for validation.

\noindent\textbf{Evaluation Metrics.} We adopt four widely-used evaluation metrics to assess image quality. 
For full-reference evaluation, we use PSNR (Peak Signal-to-Noise Ratio) and SSIM (Structural Similarity Index), which measure the content and structural similarity to the reference image, respectively. 
For no-reference evaluation, we report UIQM~\cite{panetta2015human} and UCIQE~\cite{yang2015underwater}, two perceptual quality metrics commonly used in underwater image enhancement. These metrics jointly evaluate colourfulness, contrast, sharpness, and structural fidelity, providing a comprehensive assessment of visual quality.

\begin{figure*}[t]

        \centering
        \includegraphics[width=\linewidth]{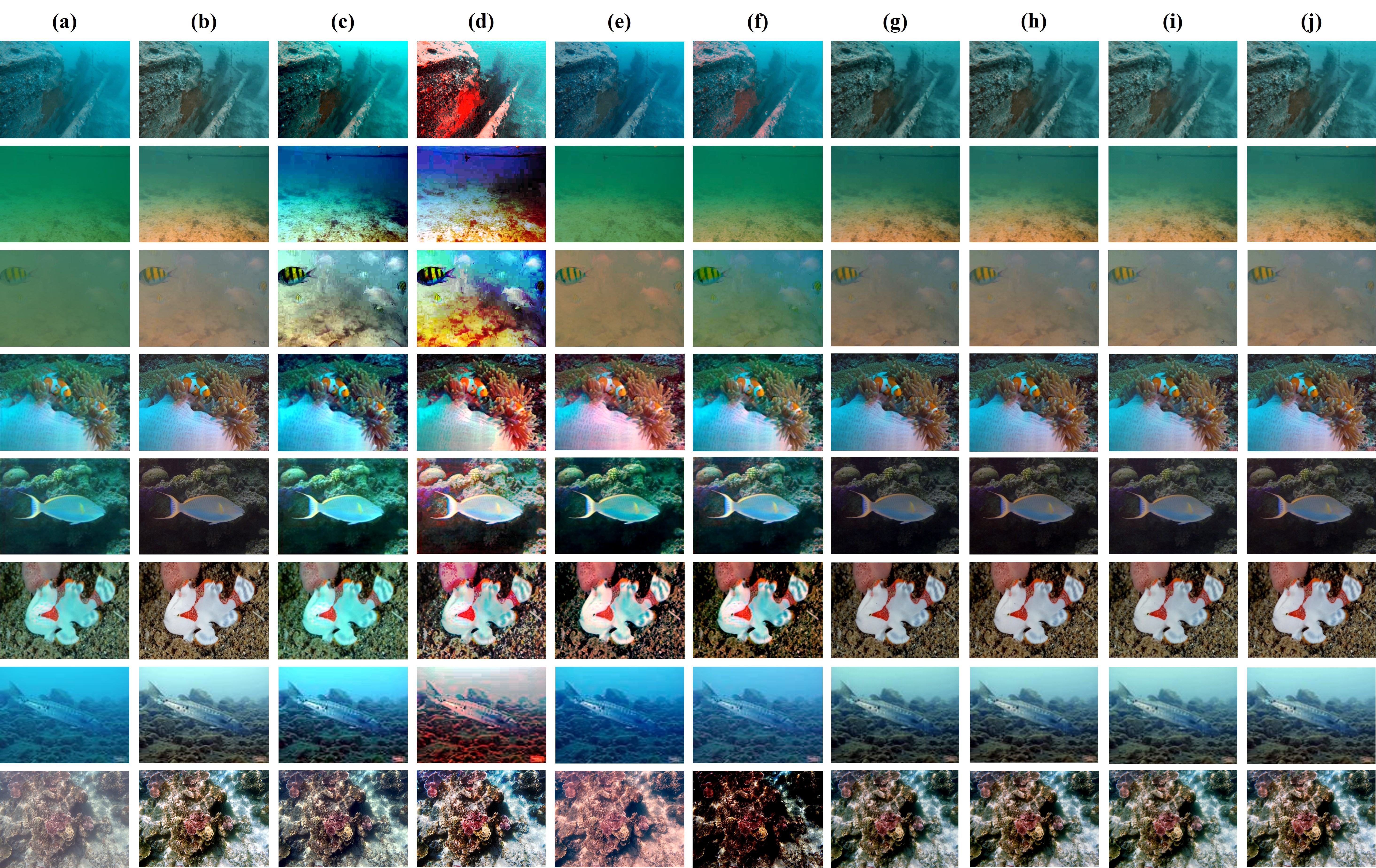 } 
    \caption{Visual comparison with different methods on real-world datasets (UIEB and LSUI). The underwater images and their corresponding enhanced images from the following methods are presented. Besides, the raw and reference images are displayed in the first two columns. (a) Raw (Input). (b) Reference (Ground Truth). (c) RGHS~\cite{huang2018shallow}. (d) RayleighDistribution~\cite{abdul2014underwater}. (e) ULAP~\cite{song2018rapid}. (f) IBLA~\cite{peng2017underwater}. (g) SMDR-IS~\cite{zhang2024synergistic}. (h) DM\_underwater~\cite{tang2023underwater}. (i)  GUPDM~\cite{mu2023generalized}. (j) DACA-Net (Ours). }
    \label{Visualization}
\end{figure*}

\subsection{Baseline Methods}
\label{4.2}

We compare our proposed method with a variety of representative UIE approaches, including physics-based and deep learning-based methods. 
Specifically, the physics-based methods include RGHS~\cite{huang2018shallow}, RayleighDistribution~\cite{abdul2014underwater}, IBLA~\cite{peng2017underwater}, and ULAP~\cite{song2018rapid}, which rely on handcrafted priors such as illumination, haze removal, or wavelength compensation.
For learning-based baselines, we consider SMDR-IS~\cite{zhang2024synergistic}, GUPDM~\cite{mu2023generalized} and DM\_underwater~\cite{tang2023underwater}, which represent recent advances in data-driven underwater enhancement. 

\subsection{Result and Analysis}
\label{4.3}

\subsubsection{Degradation Estimator Performance Comparison}
We first evaluate the accuracy and efficiency of our proposed lightweight degradation estimator by comparing it with several widely used CNN architectures.
As shown in~\Cref{DegradationEstimator}, our model achieves the best performance, obtaining the lowest MSE (0.0012) and MAE (0.0271), as well as the highest Pearson correlation coefficient (96.03\%). Moreover, our estimator demonstrates superior efficiency, using significantly fewer parameters (1.02G), lower computational cost (0.27 GFLOPs), and faster inference time (1.75 ms). 
These results indicate that our lightweight estimator precisely captures degradation information with excellent computational efficiency, validating its effectiveness for degradation-aware conditional restoration.

\subsubsection{Comparison of denoising networks}

\Cref{denoising} compares our proposed model with three representative denoising networks (UNet, ViT, and TransUNet) in terms of parameters, computational complexity (GFLOPs), inference speed, and image enhancement quality (PSNR and SSIM). 
Our method achieves the highest PSNR (28.60 dB) and SSIM (0.9456), significantly outperforming all baseline networks. Although our approach has a larger model size (88.60M parameters) and moderate computational complexity (22.27 GFLOPs), its inference speed remains efficient (3.02 ms). 
These results confirm that our model effectively balances computational cost and performance, delivering superior visual enhancement quality.

\begin{table}[t]
\caption{Comparison of denoising networks in terms of Parameters (M), FLOPs (G), Inference Time (ms), PSNR, and SSIM.}
\centering
\begin{tabular}{c|c|c|c|c|c}
\hline
\textbf{Network} & \textbf{Param.} & \textbf{FLOPs} & \textbf{Time} & \textbf{PSNR} & \textbf{SSIM} \\
\hline
UNet & 7.77 & 41.83 & 2.51 & 20.53 & 0.8803  \\
ViT & 97.81 & 31.22 & 13.28 & 24.42 & 0.9248 \\
TransUNet & 11.05 & 4.60 & 2.51 & 23.04 & 0.8983 \\
Ours & 88.60 & 22.27 & 3.02 & 28.60 & 0.9456 \\
\hline
\end{tabular}
\label{denoising}
\end{table}

\subsubsection{Quantitative comparison on the UIEB and LSUI datasets with references. }
\Cref{Quantitative} presents the quantitative evaluation results of DACA-Net and several state-of-the-art underwater image enhancement methods, including both physics-based and deep-learning-based approaches, on the UIEB and LSUI datasets. Evaluation metrics include full-reference measures (PSNR, SSIM) and no-reference perceptual quality scores (UIQM, UCIQE).
Across both datasets, DACA-Net achieves the best overall performance, consistently outperforming all competing methods in terms of colour fidelity, structural similarity, and visual quality. Specifically, on the UIEB dataset, DACA-Net achieves a PSNR of 28.60 dB and SSIM of 0.9456, outperforming the previous best GUPDM by 0.67 dB and 0.0102, respectively. In terms of perceptual scores, it also obtains the highest UIQM (3.0154) and the highest UCIQE (0.7138), indicating superior visual consistency and contrast enhancement.
On the more challenging LSUI dataset, DACA-Net further demonstrates its robustness by achieving the highest PSNR of 28.57 dB and SSIM of 0.9344, with an improvement of 1.15 dB in PSNR and 0.0093 in SSIM over GUPDM. It also ranks first in UIQM and UCIQE, scoring 2.9136 and 0.7183, respectively. Compared to classical physics-based methods (e.g., RGHS, IBLA) and representative Deep Learning-based baselines (e.g., WaterNet, SMDR-IS), our method exhibits significant advantages in both objective and subjective quality metrics.
These results highlight the effectiveness of the proposed degradation-aware conditional framework, particularly its ability to generalise across datasets and improve restoration quality under varying underwater degradation conditions.

\begin{table*}[h]
\centering

\caption{Quantitative comparison of different underwater image enhancement methods on the UIEB and LSUI datasets using four metrics: PSNR, SSIM, UIQM, and UCIQE. Compared with both physics-based and deep-learning-based baselines, the proposed DACA-Net consistently achieves the best or second-best performance across all metrics and datasets. The best results are highlighted in bold, and the second-best results are underlined.}
\begin{tabular}{c|c|cccc|cccc}
\hline
\multirow{2}{*}{\textbf{Type}} &  
\multirow{2}{*}{\textbf{Method}} &  
\multicolumn{4}{c|}{\textbf{UIEB}} & 
\multicolumn{4}{c}{\textbf{LSUI}} \\
\cline{3-10}
 & & \textbf{PSNR} & \textbf{SSIM} & \textbf{UIQM} & \textbf{UCIQE} & \textbf{PSNR} & \textbf{SSIM} & \textbf{UIQM} & \textbf{UCIQE} \\
\hline
\multirow{4}{*}{Physics-based} 
& RGHS~\cite{huang2018shallow}      & 16.02     & 0.6881 & 1.7521 & 0.6235 & 16.93    & 0.6713   & 2.0510 & 0.6573   \\
& RayleighDistribution~\cite{abdul2014underwater}      & 13.21   & 0.6300 & 2.4842 & 0.6124 & 13.57    & 0.5503    & 2.5348 & 0.6423   \\
& IBLA~\cite{peng2017underwater}       & 19.75     & 0.8193 & 1.7111 & 0.5912 & 19.43    & 0.7259    & 1.9891 & 0.6068    \\
& ULAP~\cite{song2018rapid}       & 16.60   & 0.7360 & 1.4050 & 0.5897 & 18.60    & 0.6119    & 1.5546 & 0.6022    \\
\hline
\multirow{4}{*}{DL-based}  
& SMDR-IS~\cite{zhang2024synergistic}       & 23.65   & 0.9128 & 2.9166 & 0.6079 & 22.31    & 0.8863    & 2.8173 & 0.6124   \\
& DM\_underwater~\cite{tang2023underwater}      & 27.93   & 0.9354 & \underline{2.9785} & \textbf{0.7230} & 26.12    &  0.8911    & \underline{2.8671} &  0.6538  \\
& GUPDM~\cite{mu2023generalized}         & \underline{28.19}   & \underline{0.9398} & 2.9546 & 0.7021 & \underline{27.42}   & \underline{0.9247}    & 2.8469 & \underline{0.7066} \\
& \textbf{DACA-Net (Ours)} & \textbf{28.60}   & \textbf{0.9456} & \textbf{3.0154} & \underline{0.7138} & \textbf{28.57}    & \textbf{0.9344}    & \textbf{2.9136} & \textbf{0.7183} \\
\hline
\end{tabular}%
\label{Quantitative}

\end{table*}

\subsubsection{Qualitative comparison.}
\Cref{Visualization} presents the qualitative comparison results of our proposed DACA-Net against state-of-the-art methods on real-world underwater datasets.
It can be observed that traditional physics-based methods often suffer from colour distortion or over-enhancement in complex underwater scenes, failing to maintain a balance between colour restoration and structure preservation. 
Learning-based methods show better performance in denoising and enhancement, yet some still exhibit artifacts such as blurry textures or inconsistent tones.
In contrast, our DACA-Net consistently produces visually pleasing results across diverse scenes, achieving more accurate colour fidelity, better detail retention, and effective noise suppression, demonstrating the superior capability of our method.

\begin{table}[t]
\caption{Ablation study on the effectiveness of PGFM and AdaGN modules of DACA-Net on UIEB dataset.}
\centering
\begin{tabular}{cc|cc}
\hline
\multicolumn{2}{c|}{\textbf{Module}} &  \multicolumn{2}{c}{\textbf{Metrics}}  \\
\hline
\textbf{PGFM} & \textbf{AdaGN}  & \textbf{PSNR} & \textbf{SSIM} \\
\hline
- & - & 23.72 & 0.8885  \\
 \usym{1F5F8} & - & 25.11 & 0.9010    \\
- &  \usym{1F5F8} & 27.04 & 0.9217   \\
 \usym{1F5F8} &  \usym{1F5F8} & \textbf{28.60} & \textbf{0.9456} \\
\hline
\end{tabular}
\label{Module}
\end{table}

\begin{table}[t]
\caption{ Ablation study of individual loss components in our hybrid loss design on UIEB dataset.}
\centering
\begin{tabular}{ccc|cc}
\hline
\multicolumn{3}{c|}{\textbf{Component}} &  \multicolumn{2}{c}{\textbf{Metrics}}  \\
\hline
\textbf{$\mathbf{L_{hist}}$} & \textbf{$\mathbf{L_{perc}}$}  & \textbf{$\mathbf{L_{contra}}$}  & \textbf{PSNR} & \textbf{SSIM} \\
\hline
 \usym{1F5F8} & - & - & 20.52 & 0.7698   \\
 - & \usym{1F5F8} & - & 20.44 & 0.7751   \\
 - & - & \usym{1F5F8} & 21.15 & 0.8024   \\
 \usym{1F5F8} & \usym{1F5F8} & - & 17.76 & 0.6947   \\
 \usym{1F5F8} & - & \usym{1F5F8} & 23.20 & 0.8711   \\
 - & \usym{1F5F8} & \usym{1F5F8} & 22.14 & 0.8373   \\
 \usym{1F5F8} &  \usym{1F5F8} &  \usym{1F5F8} & \textbf{28.60} & \textbf{0.9456} \\
\hline
\end{tabular}
\label{LOSS}
\end{table}

\begin{figure}[t]

        \centering
        \includegraphics[width=0.95\linewidth]{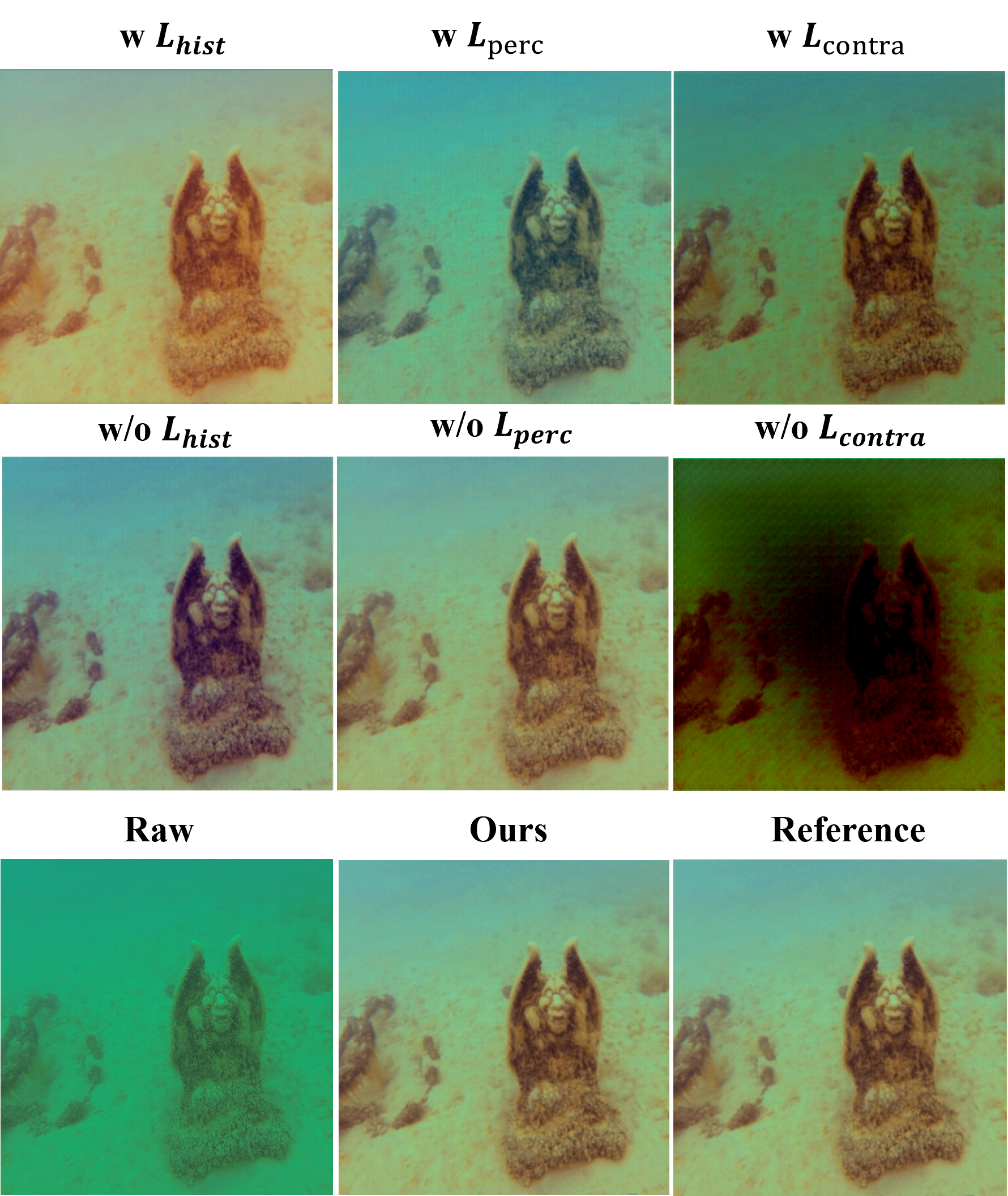 } 
    \caption{Visualisation for ablation study of loss function.}
    \label{Ablation_loss}
\end{figure}

\subsection{Ablation Study}
\label{4.4}

\subsubsection{Study on Loss Function.}
To evaluate the effectiveness of each component in our hybrid loss design, we conduct comprehensive ablation studies on the UIEB dataset. The quantitative results are summarised in~\Cref{LOSS}, while~\Cref{Ablation_loss} presents the corresponding visual comparisons, where we select a representative example with clearly visible artifacts for illustration.
We investigate three loss components: histogram alignment loss ($\mathbf{L_{hist}}$), perceptual loss ($\mathbf{L_{perc}}$), and contrastive loss ($\mathbf{L_{contra}}$). 
When all three are removed, we adopt a basic MSE loss as a baseline. This configuration yields the lowest PSNR (15.08) and SSIM (0.6789), indicating poor perceptual and structural fidelity.
From the visual results, removing $\mathbf{L_{perc}}$ leads to significant blurring and loss of texture details, while excluding $\mathbf{L_{contra}}$ causes severe colour artifacts and reduced semantic consistency. Without $\mathbf{L_{hist}}$, the model exhibits strong colour shifts as it fails to align global colour distributions.
In contrast, combining all three components results in the highest performance (PSNR: 28.60, SSIM: 0.9456) and the best visual quality, achieving effective noise removal, vivid colour restoration, and fine-grained detail retention. These results confirm the complementary roles and necessity of each component in enhancing underwater imagery.

\subsubsection{Study on Adaptive Group Normalisation and Physical-guided Fusion Module.}
We further analyse the effectiveness of the two key modules—PGFM and AdaGN—through ablation experiments, with quantitative results summarised in~\Cref{Module}. 
Incorporating only AdaGN significantly improves performance, achieving a PSNR of 27.04 dB and an SSIM of 0.9217, indicating its strong capability in adaptively guiding feature normalisation. 
Using only PGFM yields moderate improvements (PSNR=25.11 dB, SSIM=0.9010), highlighting its contribution to feature fusion enhancement. Removing both modules causes substantial performance degradation. 
These results demonstrate the complementary importance and necessity of integrating both PGFM and AdaGN in our proposed DACA-Net.

\section{Conclusion}
In this paper, we propose DACA-Net, a degradation-aware conditional diffusion model for robust and adaptive UIE. 
Specifically, we introduced a lightweight CNN-based degradation estimator to provide semantic conditioning signals that adaptively guide the denoising process. 
We further designed a novel conditional diffusion framework utilising a Swin UNet backbone, a degradation-guided adaptive normalisation module, and a physical-guided feature fusion module to incorporate underwater imaging priors effectively. 
A hybrid loss function was introduced to further improve restoration fidelity. 
Comprehensive experimental results on standard underwater datasets demonstrated that our method outperforms state-of-the-art approaches in terms of both quantitative metrics and visual quality, verifying its effectiveness and robustness for real-world underwater scenarios.

\section{Acknowledgements}
This work is supported by Guangdong Provincial Key Lab of Integrated Communication, Sensing and
Computation for Ubiquitous Internet of Things (No.2023B1212010007), China NSFC Grant award number (No.62472366), 111 Center (No.D25008), the Project of DEGP (No.2024GCZX003, 2023KCXTD042), Shenzhen Science and Technology Foundation (ZDSYS20190902092853047) and The Hong Kong Branch of the Southern Marine Science and Engineering Guangdong Laboratory (Guangzhou).

\bibliographystyle{ACM-Reference-Format}
\bibliography{reference}


\begin{thebibliography}{65}


\ifx \showCODEN    \undefined \def \showCODEN     #1{\unskip}     \fi
\ifx \showDOI      \undefined \def \showDOI       #1{#1}\fi
\ifx \showISBNx    \undefined \def \showISBNx     #1{\unskip}     \fi
\ifx \showISBNxiii \undefined \def \showISBNxiii  #1{\unskip}     \fi
\ifx \showISSN     \undefined \def \showISSN      #1{\unskip}     \fi
\ifx \showLCCN     \undefined \def \showLCCN      #1{\unskip}     \fi
\ifx \shownote     \undefined \def \shownote      #1{#1}          \fi
\ifx \showarticletitle \undefined \def \showarticletitle #1{#1}   \fi
\ifx \showURL      \undefined \def \showURL       {\relax}        \fi
\providecommand\bibfield[2]{#2}
\providecommand\bibinfo[2]{#2}
\providecommand\natexlab[1]{#1}
\providecommand\showeprint[2][]{arXiv:#2}

\bibitem[Abdul~Ghani and Mat~Isa(2014)]%
        {abdul2014underwater}
\bibfield{author}{\bibinfo{person}{Ahmad~Shahrizan Abdul~Ghani} {and} \bibinfo{person}{Nor~Ashidi Mat~Isa}.} \bibinfo{year}{2014}\natexlab{}.
\newblock \showarticletitle{Underwater image quality enhancement through composition of dual-intensity images and Rayleigh-stretching}.
\newblock \bibinfo{journal}{\emph{SpringerPlus}}  \bibinfo{volume}{3} (\bibinfo{year}{2014}), \bibinfo{pages}{1--14}.
\newblock


\bibitem[Afifi and Brown(2019)]%
        {afifi2019else}
\bibfield{author}{\bibinfo{person}{Mahmoud Afifi} {and} \bibinfo{person}{Michael~S Brown}.} \bibinfo{year}{2019}\natexlab{}.
\newblock \showarticletitle{What else can fool deep learning? Addressing color constancy errors on deep neural network performance}. In \bibinfo{booktitle}{\emph{Proceedings of the IEEE/CVF international conference on computer vision}}. \bibinfo{pages}{243--252}.
\newblock


\bibitem[Akkaynak and Treibitz(2018)]%
        {akkaynak2018revised}
\bibfield{author}{\bibinfo{person}{Derya Akkaynak} {and} \bibinfo{person}{Tali Treibitz}.} \bibinfo{year}{2018}\natexlab{}.
\newblock \showarticletitle{A revised underwater image formation model}. In \bibinfo{booktitle}{\emph{Proceedings of the IEEE conference on computer vision and pattern recognition}}. \bibinfo{pages}{6723--6732}.
\newblock


\bibitem[Akkaynak and Treibitz(2019)]%
        {akkaynak2019sea}
\bibfield{author}{\bibinfo{person}{Derya Akkaynak} {and} \bibinfo{person}{Tali Treibitz}.} \bibinfo{year}{2019}\natexlab{}.
\newblock \showarticletitle{Sea-thru: A method for removing water from underwater images}. In \bibinfo{booktitle}{\emph{Proceedings of the IEEE/CVF conference on computer vision and pattern recognition}}. \bibinfo{pages}{1682--1691}.
\newblock


\bibitem[Brock et~al\mbox{.}(2018)]%
        {brock2018large}
\bibfield{author}{\bibinfo{person}{Andrew Brock}, \bibinfo{person}{Jeff Donahue}, {and} \bibinfo{person}{Karen Simonyan}.} \bibinfo{year}{2018}\natexlab{}.
\newblock \showarticletitle{Large scale GAN training for high fidelity natural image synthesis}.
\newblock \bibinfo{journal}{\emph{arXiv preprint arXiv:1809.11096}} (\bibinfo{year}{2018}).
\newblock


\bibitem[Cao et~al\mbox{.}(2022)]%
        {cao2022swin}
\bibfield{author}{\bibinfo{person}{Hu Cao}, \bibinfo{person}{Yueyue Wang}, \bibinfo{person}{Joy Chen}, \bibinfo{person}{Dongsheng Jiang}, \bibinfo{person}{Xiaopeng Zhang}, \bibinfo{person}{Qi Tian}, {and} \bibinfo{person}{Manning Wang}.} \bibinfo{year}{2022}\natexlab{}.
\newblock \showarticletitle{Swin-unet: Unet-like pure transformer for medical image segmentation}. In \bibinfo{booktitle}{\emph{European conference on computer vision}}. Springer, \bibinfo{pages}{205--218}.
\newblock


\bibitem[Chen et~al\mbox{.}(2020)]%
        {chen2020simple}
\bibfield{author}{\bibinfo{person}{Ting Chen}, \bibinfo{person}{Simon Kornblith}, \bibinfo{person}{Mohammad Norouzi}, {and} \bibinfo{person}{Geoffrey Hinton}.} \bibinfo{year}{2020}\natexlab{}.
\newblock \showarticletitle{A simple framework for contrastive learning of visual representations}. In \bibinfo{booktitle}{\emph{International conference on machine learning}}. PmLR, \bibinfo{pages}{1597--1607}.
\newblock


\bibitem[Chiang and Chen(2011)]%
        {chiang2011underwater}
\bibfield{author}{\bibinfo{person}{John~Y Chiang} {and} \bibinfo{person}{Ying-Ching Chen}.} \bibinfo{year}{2011}\natexlab{}.
\newblock \showarticletitle{Underwater image enhancement by wavelength compensation and dehazing}.
\newblock \bibinfo{journal}{\emph{IEEE transactions on image processing}} \bibinfo{volume}{21}, \bibinfo{number}{4} (\bibinfo{year}{2011}), \bibinfo{pages}{1756--1769}.
\newblock


\bibitem[Chollet(2017)]%
        {chollet2017xception}
\bibfield{author}{\bibinfo{person}{Fran{\c{c}}ois Chollet}.} \bibinfo{year}{2017}\natexlab{}.
\newblock \showarticletitle{Xception: Deep learning with depthwise separable convolutions}. In \bibinfo{booktitle}{\emph{Proceedings of the IEEE conference on computer vision and pattern recognition}}. \bibinfo{pages}{1251--1258}.
\newblock


\bibitem[Danovaro et~al\mbox{.}(2017)]%
        {danovaro2017deep}
\bibfield{author}{\bibinfo{person}{Roberto Danovaro}, \bibinfo{person}{Cinzia Corinaldesi}, \bibinfo{person}{Antonio Dell’Anno}, {and} \bibinfo{person}{Paul~VR Snelgrove}.} \bibinfo{year}{2017}\natexlab{}.
\newblock \showarticletitle{The deep-sea under global change}.
\newblock \bibinfo{journal}{\emph{Current Biology}} \bibinfo{volume}{27}, \bibinfo{number}{11} (\bibinfo{year}{2017}), \bibinfo{pages}{R461--R465}.
\newblock


\bibitem[Edinger and Risk(2000)]%
        {edinger2000reef}
\bibfield{author}{\bibinfo{person}{Evan~N Edinger} {and} \bibinfo{person}{Michael~J Risk}.} \bibinfo{year}{2000}\natexlab{}.
\newblock \showarticletitle{Reef classification by coral morphology predicts coral reef conservation value}.
\newblock \bibinfo{journal}{\emph{Biological Conservation}} \bibinfo{volume}{92}, \bibinfo{number}{1} (\bibinfo{year}{2000}), \bibinfo{pages}{1--13}.
\newblock


\bibitem[Fabbri et~al\mbox{.}(2018)]%
        {fabbri2018enhancing}
\bibfield{author}{\bibinfo{person}{Cameron Fabbri}, \bibinfo{person}{Md~Jahidul Islam}, {and} \bibinfo{person}{Junaed Sattar}.} \bibinfo{year}{2018}\natexlab{}.
\newblock \showarticletitle{Enhancing underwater imagery using generative adversarial networks}. In \bibinfo{booktitle}{\emph{2018 IEEE international conference on robotics and automation (ICRA)}}. IEEE, \bibinfo{pages}{7159--7165}.
\newblock


\bibitem[Ferrari et~al\mbox{.}(2016)]%
        {ferrari2016quantifying}
\bibfield{author}{\bibinfo{person}{Renata Ferrari}, \bibinfo{person}{David McKinnon}, \bibinfo{person}{Hu He}, \bibinfo{person}{Ryan~N Smith}, \bibinfo{person}{Peter Corke}, \bibinfo{person}{Manuel Gonz{\'a}lez-Rivero}, \bibinfo{person}{Peter~J Mumby}, {and} \bibinfo{person}{Ben Upcroft}.} \bibinfo{year}{2016}\natexlab{}.
\newblock \showarticletitle{Quantifying multiscale habitat structural complexity: a cost-effective framework for underwater 3D modelling}.
\newblock \bibinfo{journal}{\emph{Remote Sensing}} \bibinfo{volume}{8}, \bibinfo{number}{2} (\bibinfo{year}{2016}), \bibinfo{pages}{113}.
\newblock


\bibitem[Fu et~al\mbox{.}(2022)]%
        {fu2022unsupervised}
\bibfield{author}{\bibinfo{person}{Zhenqi Fu}, \bibinfo{person}{Huangxing Lin}, \bibinfo{person}{Yan Yang}, \bibinfo{person}{Shu Chai}, \bibinfo{person}{Liyan Sun}, \bibinfo{person}{Yue Huang}, {and} \bibinfo{person}{Xinghao Ding}.} \bibinfo{year}{2022}\natexlab{}.
\newblock \showarticletitle{Unsupervised underwater image restoration: From a homology perspective}. In \bibinfo{booktitle}{\emph{Proceedings of the AAAI conference on artificial intelligence}}, Vol.~\bibinfo{volume}{36}. \bibinfo{pages}{643--651}.
\newblock


\bibitem[Gatys et~al\mbox{.}(2015)]%
        {gatys2015neural}
\bibfield{author}{\bibinfo{person}{Leon~A Gatys}, \bibinfo{person}{Alexander~S Ecker}, {and} \bibinfo{person}{Matthias Bethge}.} \bibinfo{year}{2015}\natexlab{}.
\newblock \showarticletitle{A neural algorithm of artistic style}.
\newblock \bibinfo{journal}{\emph{arXiv preprint arXiv:1508.06576}} (\bibinfo{year}{2015}).
\newblock


\bibitem[Goodfellow et~al\mbox{.}(2014)]%
        {goodfellow2014generative}
\bibfield{author}{\bibinfo{person}{Ian~J Goodfellow}, \bibinfo{person}{Jean Pouget-Abadie}, \bibinfo{person}{Mehdi Mirza}, \bibinfo{person}{Bing Xu}, \bibinfo{person}{David Warde-Farley}, \bibinfo{person}{Sherjil Ozair}, \bibinfo{person}{Aaron Courville}, {and} \bibinfo{person}{Yoshua Bengio}.} \bibinfo{year}{2014}\natexlab{}.
\newblock \showarticletitle{Generative adversarial nets}.
\newblock \bibinfo{journal}{\emph{Advances in neural information processing systems}}  \bibinfo{volume}{27} (\bibinfo{year}{2014}).
\newblock


\bibitem[Guan et~al\mbox{.}(2023)]%
        {guan2023diffwater}
\bibfield{author}{\bibinfo{person}{Meisheng Guan}, \bibinfo{person}{Haiyong Xu}, \bibinfo{person}{Gangyi Jiang}, \bibinfo{person}{Mei Yu}, \bibinfo{person}{Yeyao Chen}, \bibinfo{person}{Ting Luo}, {and} \bibinfo{person}{Xuebo Zhang}.} \bibinfo{year}{2023}\natexlab{}.
\newblock \showarticletitle{DiffWater: Underwater image enhancement based on conditional denoising diffusion probabilistic model}.
\newblock \bibinfo{journal}{\emph{IEEE Journal of Selected Topics in Applied Earth Observations and Remote Sensing}}  \bibinfo{volume}{17} (\bibinfo{year}{2023}), \bibinfo{pages}{2319--2335}.
\newblock


\bibitem[He et~al\mbox{.}(2010)]%
        {he2010single}
\bibfield{author}{\bibinfo{person}{Kaiming He}, \bibinfo{person}{Jian Sun}, {and} \bibinfo{person}{Xiaoou Tang}.} \bibinfo{year}{2010}\natexlab{}.
\newblock \showarticletitle{Single image haze removal using dark channel prior}.
\newblock \bibinfo{journal}{\emph{IEEE transactions on pattern analysis and machine intelligence}} \bibinfo{volume}{33}, \bibinfo{number}{12} (\bibinfo{year}{2010}), \bibinfo{pages}{2341--2353}.
\newblock


\bibitem[He et~al\mbox{.}(2016)]%
        {he2016deep}
\bibfield{author}{\bibinfo{person}{Kaiming He}, \bibinfo{person}{Xiangyu Zhang}, \bibinfo{person}{Shaoqing Ren}, {and} \bibinfo{person}{Jian Sun}.} \bibinfo{year}{2016}\natexlab{}.
\newblock \showarticletitle{Deep residual learning for image recognition}. In \bibinfo{booktitle}{\emph{Proceedings of the IEEE conference on computer vision and pattern recognition}}. \bibinfo{pages}{770--778}.
\newblock


\bibitem[Ho et~al\mbox{.}(2020)]%
        {ho2020denoising}
\bibfield{author}{\bibinfo{person}{Jonathan Ho}, \bibinfo{person}{Ajay Jain}, {and} \bibinfo{person}{Pieter Abbeel}.} \bibinfo{year}{2020}\natexlab{}.
\newblock \showarticletitle{Denoising diffusion probabilistic models}.
\newblock \bibinfo{journal}{\emph{Advances in neural information processing systems}}  \bibinfo{volume}{33} (\bibinfo{year}{2020}), \bibinfo{pages}{6840--6851}.
\newblock


\bibitem[Hoegh-Guldberg et~al\mbox{.}(2023)]%
        {hoegh2023ocean}
\bibfield{author}{\bibinfo{person}{Ove Hoegh-Guldberg}, \bibinfo{person}{Ken Caldeira}, \bibinfo{person}{Thierry Chopin}, \bibinfo{person}{Steve Gaines}, \bibinfo{person}{Peter Haugan}, \bibinfo{person}{Mark Hemer}, \bibinfo{person}{Jennifer Howard}, \bibinfo{person}{Manaswita Konar}, \bibinfo{person}{Dorte Krause-Jensen}, \bibinfo{person}{Catherine~E Lovelock}, {et~al\mbox{.}}} \bibinfo{year}{2023}\natexlab{}.
\newblock \showarticletitle{The ocean as a solution to climate change: five opportunities for action}.
\newblock In \bibinfo{booktitle}{\emph{The blue compendium: From knowledge to action for a sustainable ocean economy}}. \bibinfo{publisher}{Springer}, \bibinfo{pages}{619--680}.
\newblock


\bibitem[Howard et~al\mbox{.}(2017)]%
        {howard2017mobilenets}
\bibfield{author}{\bibinfo{person}{Andrew~G Howard}, \bibinfo{person}{Menglong Zhu}, \bibinfo{person}{Bo Chen}, \bibinfo{person}{Dmitry Kalenichenko}, \bibinfo{person}{Weijun Wang}, \bibinfo{person}{Tobias Weyand}, \bibinfo{person}{Marco Andreetto}, {and} \bibinfo{person}{Hartwig Adam}.} \bibinfo{year}{2017}\natexlab{}.
\newblock \showarticletitle{Mobilenets: Efficient convolutional neural networks for mobile vision applications}.
\newblock \bibinfo{journal}{\emph{arXiv preprint arXiv:1704.04861}} (\bibinfo{year}{2017}).
\newblock


\bibitem[Huang et~al\mbox{.}(2018)]%
        {huang2018shallow}
\bibfield{author}{\bibinfo{person}{Dongmei Huang}, \bibinfo{person}{Yan Wang}, \bibinfo{person}{Wei Song}, \bibinfo{person}{Jean Sequeira}, {and} \bibinfo{person}{S{\'e}bastien Mavromatis}.} \bibinfo{year}{2018}\natexlab{}.
\newblock \showarticletitle{Shallow-water image enhancement using relative global histogram stretching based on adaptive parameter acquisition}. In \bibinfo{booktitle}{\emph{MultiMedia Modeling: 24th International Conference, MMM 2018, Bangkok, Thailand, February 5-7, 2018, Proceedings, Part I 24}}. Springer, \bibinfo{pages}{453--465}.
\newblock


\bibitem[Hughes et~al\mbox{.}(2017)]%
        {hughes2017global}
\bibfield{author}{\bibinfo{person}{Terry~P Hughes}, \bibinfo{person}{James~T Kerry}, \bibinfo{person}{Mariana {\'A}lvarez-Noriega}, \bibinfo{person}{Jorge~G {\'A}lvarez-Romero}, \bibinfo{person}{Kristen~D Anderson}, \bibinfo{person}{Andrew~H Baird}, \bibinfo{person}{Russell~C Babcock}, \bibinfo{person}{Maria Beger}, \bibinfo{person}{David~R Bellwood}, \bibinfo{person}{Ray Berkelmans}, {et~al\mbox{.}}} \bibinfo{year}{2017}\natexlab{}.
\newblock \showarticletitle{Global warming and recurrent mass bleaching of corals}.
\newblock \bibinfo{journal}{\emph{Nature}} \bibinfo{volume}{543}, \bibinfo{number}{7645} (\bibinfo{year}{2017}), \bibinfo{pages}{373--377}.
\newblock


\bibitem[Jerlov(1976)]%
        {jerlov1976marine}
\bibfield{author}{\bibinfo{person}{Nils~Gunnar Jerlov}.} \bibinfo{year}{1976}\natexlab{}.
\newblock \bibinfo{booktitle}{\emph{Marine optics}}. Vol.~\bibinfo{volume}{14}.
\newblock \bibinfo{publisher}{Elsevier}.
\newblock


\bibitem[Jobson et~al\mbox{.}(1997)]%
        {jobson1997multiscale}
\bibfield{author}{\bibinfo{person}{Daniel~J Jobson}, \bibinfo{person}{Zia-ur Rahman}, {and} \bibinfo{person}{Glenn~A Woodell}.} \bibinfo{year}{1997}\natexlab{}.
\newblock \showarticletitle{A multiscale retinex for bridging the gap between color images and the human observation of scenes}.
\newblock \bibinfo{journal}{\emph{IEEE Transactions on Image processing}} \bibinfo{volume}{6}, \bibinfo{number}{7} (\bibinfo{year}{1997}), \bibinfo{pages}{965--976}.
\newblock


\bibitem[Johnson et~al\mbox{.}(2016)]%
        {johnson2016perceptual}
\bibfield{author}{\bibinfo{person}{Justin Johnson}, \bibinfo{person}{Alexandre Alahi}, {and} \bibinfo{person}{Li Fei-Fei}.} \bibinfo{year}{2016}\natexlab{}.
\newblock \showarticletitle{Perceptual losses for real-time style transfer and super-resolution}. In \bibinfo{booktitle}{\emph{Computer Vision--ECCV 2016: 14th European Conference, Amsterdam, The Netherlands, October 11-14, 2016, Proceedings, Part II 14}}. Springer, \bibinfo{pages}{694--711}.
\newblock


\bibitem[Kirk(1994)]%
        {kirk1994light}
\bibfield{author}{\bibinfo{person}{John~TO Kirk}.} \bibinfo{year}{1994}\natexlab{}.
\newblock \bibinfo{booktitle}{\emph{Light and photosynthesis in aquatic ecosystems}}.
\newblock \bibinfo{publisher}{Cambridge university press}.
\newblock


\bibitem[Krizhevsky et~al\mbox{.}(2012)]%
        {krizhevsky2012imagenet}
\bibfield{author}{\bibinfo{person}{Alex Krizhevsky}, \bibinfo{person}{Ilya Sutskever}, {and} \bibinfo{person}{Geoffrey~E Hinton}.} \bibinfo{year}{2012}\natexlab{}.
\newblock \showarticletitle{Imagenet classification with deep convolutional neural networks}.
\newblock \bibinfo{journal}{\emph{Advances in neural information processing systems}}  \bibinfo{volume}{25} (\bibinfo{year}{2012}).
\newblock


\bibitem[Kulikov et~al\mbox{.}(2023)]%
        {kulikov2023sinddm}
\bibfield{author}{\bibinfo{person}{Vladimir Kulikov}, \bibinfo{person}{Shahar Yadin}, \bibinfo{person}{Matan Kleiner}, {and} \bibinfo{person}{Tomer Michaeli}.} \bibinfo{year}{2023}\natexlab{}.
\newblock \showarticletitle{Sinddm: A single image denoising diffusion model}. In \bibinfo{booktitle}{\emph{International conference on machine learning}}. PMLR, \bibinfo{pages}{17920--17930}.
\newblock


\bibitem[Levin et~al\mbox{.}(2007)]%
        {levin2007closed}
\bibfield{author}{\bibinfo{person}{Anat Levin}, \bibinfo{person}{Dani Lischinski}, {and} \bibinfo{person}{Yair Weiss}.} \bibinfo{year}{2007}\natexlab{}.
\newblock \showarticletitle{A closed-form solution to natural image matting}.
\newblock \bibinfo{journal}{\emph{IEEE transactions on pattern analysis and machine intelligence}} \bibinfo{volume}{30}, \bibinfo{number}{2} (\bibinfo{year}{2007}), \bibinfo{pages}{228--242}.
\newblock


\bibitem[Li et~al\mbox{.}(2021)]%
        {li2021underwater}
\bibfield{author}{\bibinfo{person}{Chongyi Li}, \bibinfo{person}{Saeed Anwar}, \bibinfo{person}{Junhui Hou}, \bibinfo{person}{Runmin Cong}, \bibinfo{person}{Chunle Guo}, {and} \bibinfo{person}{Wenqi Ren}.} \bibinfo{year}{2021}\natexlab{}.
\newblock \showarticletitle{Underwater image enhancement via medium transmission-guided multi-color space embedding}.
\newblock \bibinfo{journal}{\emph{IEEE Transactions on Image Processing}}  \bibinfo{volume}{30} (\bibinfo{year}{2021}), \bibinfo{pages}{4985--5000}.
\newblock


\bibitem[Li et~al\mbox{.}(2020)]%
        {li2020underwater}
\bibfield{author}{\bibinfo{person}{Chongyi Li}, \bibinfo{person}{Saeed Anwar}, {and} \bibinfo{person}{Fatih Porikli}.} \bibinfo{year}{2020}\natexlab{}.
\newblock \showarticletitle{Underwater scene prior inspired deep underwater image and video enhancement}.
\newblock \bibinfo{journal}{\emph{Pattern recognition}}  \bibinfo{volume}{98} (\bibinfo{year}{2020}), \bibinfo{pages}{107038}.
\newblock


\bibitem[Li et~al\mbox{.}(2019)]%
        {li2019underwater}
\bibfield{author}{\bibinfo{person}{Chongyi Li}, \bibinfo{person}{Chunle Guo}, \bibinfo{person}{Wenqi Ren}, \bibinfo{person}{Runmin Cong}, \bibinfo{person}{Junhui Hou}, \bibinfo{person}{Sam Kwong}, {and} \bibinfo{person}{Dacheng Tao}.} \bibinfo{year}{2019}\natexlab{}.
\newblock \showarticletitle{An underwater image enhancement benchmark dataset and beyond}.
\newblock \bibinfo{journal}{\emph{IEEE transactions on image processing}}  \bibinfo{volume}{29} (\bibinfo{year}{2019}), \bibinfo{pages}{4376--4389}.
\newblock


\bibitem[Li and Guo(2015)]%
        {li2015underwater}
\bibfield{author}{\bibinfo{person}{Chongyi Li} {and} \bibinfo{person}{Jichang Guo}.} \bibinfo{year}{2015}\natexlab{}.
\newblock \showarticletitle{Underwater image enhancement by dehazing and color correction}.
\newblock \bibinfo{journal}{\emph{Journal of Electronic Imaging}} \bibinfo{volume}{24}, \bibinfo{number}{3} (\bibinfo{year}{2015}), \bibinfo{pages}{033023--033023}.
\newblock


\bibitem[Li et~al\mbox{.}(2017)]%
        {li2017watergan}
\bibfield{author}{\bibinfo{person}{Jie Li}, \bibinfo{person}{Katherine~A Skinner}, \bibinfo{person}{Ryan~M Eustice}, {and} \bibinfo{person}{Matthew Johnson-Roberson}.} \bibinfo{year}{2017}\natexlab{}.
\newblock \showarticletitle{WaterGAN: Unsupervised generative network to enable real-time color correction of monocular underwater images}.
\newblock \bibinfo{journal}{\emph{IEEE Robotics and Automation letters}} \bibinfo{volume}{3}, \bibinfo{number}{1} (\bibinfo{year}{2017}), \bibinfo{pages}{387--394}.
\newblock


\bibitem[Liu et~al\mbox{.}(2023)]%
        {liu2023fair}
\bibfield{author}{\bibinfo{person}{Tongkun Liu}, \bibinfo{person}{Bing Li}, \bibinfo{person}{Xiao Du}, \bibinfo{person}{Bingke Jiang}, \bibinfo{person}{Leqi Geng}, \bibinfo{person}{Feiyang Wang}, {and} \bibinfo{person}{Zhuo Zhao}.} \bibinfo{year}{2023}\natexlab{}.
\newblock \showarticletitle{Fair: Frequency-aware image restoration for industrial visual anomaly detection}.
\newblock \bibinfo{journal}{\emph{arXiv preprint arXiv:2309.07068}} (\bibinfo{year}{2023}).
\newblock


\bibitem[Liu et~al\mbox{.}(2021)]%
        {liu2021swin}
\bibfield{author}{\bibinfo{person}{Ze Liu}, \bibinfo{person}{Yutong Lin}, \bibinfo{person}{Yue Cao}, \bibinfo{person}{Han Hu}, \bibinfo{person}{Yixuan Wei}, \bibinfo{person}{Zheng Zhang}, \bibinfo{person}{Stephen Lin}, {and} \bibinfo{person}{Baining Guo}.} \bibinfo{year}{2021}\natexlab{}.
\newblock \showarticletitle{Swin transformer: Hierarchical vision transformer using shifted windows}. In \bibinfo{booktitle}{\emph{Proceedings of the IEEE/CVF international conference on computer vision}}. \bibinfo{pages}{10012--10022}.
\newblock


\bibitem[Lugmayr et~al\mbox{.}(2022)]%
        {lugmayr2022repaint}
\bibfield{author}{\bibinfo{person}{Andreas Lugmayr}, \bibinfo{person}{Martin Danelljan}, \bibinfo{person}{Andres Romero}, \bibinfo{person}{Fisher Yu}, \bibinfo{person}{Radu Timofte}, {and} \bibinfo{person}{Luc Van~Gool}.} \bibinfo{year}{2022}\natexlab{}.
\newblock \showarticletitle{Repaint: Inpainting using denoising diffusion probabilistic models}. In \bibinfo{booktitle}{\emph{Proceedings of the IEEE/CVF conference on computer vision and pattern recognition}}. \bibinfo{pages}{11461--11471}.
\newblock


\bibitem[Mallet and Pelletier(2014)]%
        {mallet2014underwater}
\bibfield{author}{\bibinfo{person}{Delphine Mallet} {and} \bibinfo{person}{Dominique Pelletier}.} \bibinfo{year}{2014}\natexlab{}.
\newblock \showarticletitle{Underwater video techniques for observing coastal marine biodiversity: a review of sixty years of publications (1952--2012)}.
\newblock \bibinfo{journal}{\emph{Fisheries Research}}  \bibinfo{volume}{154} (\bibinfo{year}{2014}), \bibinfo{pages}{44--62}.
\newblock


\bibitem[Moser et~al\mbox{.}(2024)]%
        {moser2024diffusion}
\bibfield{author}{\bibinfo{person}{Brian~B Moser}, \bibinfo{person}{Arundhati~S Shanbhag}, \bibinfo{person}{Federico Raue}, \bibinfo{person}{Stanislav Frolov}, \bibinfo{person}{Sebastian Palacio}, {and} \bibinfo{person}{Andreas Dengel}.} \bibinfo{year}{2024}\natexlab{}.
\newblock \showarticletitle{Diffusion models, image super-resolution, and everything: A survey}.
\newblock \bibinfo{journal}{\emph{IEEE Transactions on Neural Networks and Learning Systems}} (\bibinfo{year}{2024}).
\newblock


\bibitem[Mu et~al\mbox{.}(2023)]%
        {mu2023generalized}
\bibfield{author}{\bibinfo{person}{Pan Mu}, \bibinfo{person}{Hanning Xu}, \bibinfo{person}{Zheyuan Liu}, \bibinfo{person}{Zheng Wang}, \bibinfo{person}{Sixian Chan}, {and} \bibinfo{person}{Cong Bai}.} \bibinfo{year}{2023}\natexlab{}.
\newblock \showarticletitle{A generalized physical-knowledge-guided dynamic model for underwater image enhancement}. In \bibinfo{booktitle}{\emph{Proceedings of the 31st ACM international conference on multimedia}}. \bibinfo{pages}{7111--7120}.
\newblock


\bibitem[Obura et~al\mbox{.}(2019)]%
        {obura2019coral}
\bibfield{author}{\bibinfo{person}{David~O Obura}, \bibinfo{person}{Greta Aeby}, \bibinfo{person}{Natchanon Amornthammarong}, \bibinfo{person}{Ward Appeltans}, \bibinfo{person}{Nicholas Bax}, \bibinfo{person}{Joe Bishop}, \bibinfo{person}{Russell~E Brainard}, \bibinfo{person}{Samuel Chan}, \bibinfo{person}{Pamela Fletcher}, \bibinfo{person}{Timothy~AC Gordon}, {et~al\mbox{.}}} \bibinfo{year}{2019}\natexlab{}.
\newblock \showarticletitle{Coral reef monitoring, reef assessment technologies, and ecosystem-based management}.
\newblock \bibinfo{journal}{\emph{Frontiers in Marine Science}}  \bibinfo{volume}{6} (\bibinfo{year}{2019}), \bibinfo{pages}{580}.
\newblock


\bibitem[Panetta et~al\mbox{.}(2015)]%
        {panetta2015human}
\bibfield{author}{\bibinfo{person}{Karen Panetta}, \bibinfo{person}{Chen Gao}, {and} \bibinfo{person}{Sos Agaian}.} \bibinfo{year}{2015}\natexlab{}.
\newblock \showarticletitle{Human-visual-system-inspired underwater image quality measures}.
\newblock \bibinfo{journal}{\emph{IEEE Journal of Oceanic Engineering}} \bibinfo{volume}{41}, \bibinfo{number}{3} (\bibinfo{year}{2015}), \bibinfo{pages}{541--551}.
\newblock


\bibitem[Peng et~al\mbox{.}(2023)]%
        {peng2023u}
\bibfield{author}{\bibinfo{person}{Lintao Peng}, \bibinfo{person}{Chunli Zhu}, {and} \bibinfo{person}{Liheng Bian}.} \bibinfo{year}{2023}\natexlab{}.
\newblock \showarticletitle{U-shape transformer for underwater image enhancement}.
\newblock \bibinfo{journal}{\emph{IEEE Transactions on Image Processing}}  \bibinfo{volume}{32} (\bibinfo{year}{2023}), \bibinfo{pages}{3066--3079}.
\newblock


\bibitem[Peng and Cosman(2017)]%
        {peng2017underwater}
\bibfield{author}{\bibinfo{person}{Yan-Tsung Peng} {and} \bibinfo{person}{Pamela~C Cosman}.} \bibinfo{year}{2017}\natexlab{}.
\newblock \showarticletitle{Underwater image restoration based on image blurriness and light absorption}.
\newblock \bibinfo{journal}{\emph{IEEE transactions on image processing}} \bibinfo{volume}{26}, \bibinfo{number}{4} (\bibinfo{year}{2017}), \bibinfo{pages}{1579--1594}.
\newblock


\bibitem[Pope and Fry(1997)]%
        {pope1997absorption}
\bibfield{author}{\bibinfo{person}{Robin~M Pope} {and} \bibinfo{person}{Edward~S Fry}.} \bibinfo{year}{1997}\natexlab{}.
\newblock \showarticletitle{Absorption spectrum (380--700 nm) of pure water. II. Integrating cavity measurements}.
\newblock \bibinfo{journal}{\emph{Applied optics}} \bibinfo{volume}{36}, \bibinfo{number}{33} (\bibinfo{year}{1997}), \bibinfo{pages}{8710--8723}.
\newblock


\bibitem[Raveendran et~al\mbox{.}(2021)]%
        {raveendran2021underwater}
\bibfield{author}{\bibinfo{person}{Smitha Raveendran}, \bibinfo{person}{Mukesh~D Patil}, {and} \bibinfo{person}{Gajanan~K Birajdar}.} \bibinfo{year}{2021}\natexlab{}.
\newblock \showarticletitle{Underwater image enhancement: a comprehensive review, recent trends, challenges and applications}.
\newblock \bibinfo{journal}{\emph{Artificial Intelligence Review}}  \bibinfo{volume}{54} (\bibinfo{year}{2021}), \bibinfo{pages}{5413--5467}.
\newblock


\bibitem[Reinhard et~al\mbox{.}(2001)]%
        {reinhard2001color}
\bibfield{author}{\bibinfo{person}{Erik Reinhard}, \bibinfo{person}{Michael Adhikhmin}, \bibinfo{person}{Bruce Gooch}, {and} \bibinfo{person}{Peter Shirley}.} \bibinfo{year}{2001}\natexlab{}.
\newblock \showarticletitle{Color transfer between images}.
\newblock \bibinfo{journal}{\emph{IEEE Computer graphics and applications}} \bibinfo{volume}{21}, \bibinfo{number}{5} (\bibinfo{year}{2001}), \bibinfo{pages}{34--41}.
\newblock


\bibitem[Sahoo et~al\mbox{.}(2024)]%
        {sahoo2024diffusion}
\bibfield{author}{\bibinfo{person}{Subham Sahoo}, \bibinfo{person}{Aaron Gokaslan}, \bibinfo{person}{Christopher~M De~Sa}, {and} \bibinfo{person}{Volodymyr Kuleshov}.} \bibinfo{year}{2024}\natexlab{}.
\newblock \showarticletitle{Diffusion models with learned adaptive noise}.
\newblock \bibinfo{journal}{\emph{Advances in Neural Information Processing Systems}}  \bibinfo{volume}{37} (\bibinfo{year}{2024}), \bibinfo{pages}{105730--105779}.
\newblock


\bibitem[Simonyan and Zisserman(2014)]%
        {simonyan2014very}
\bibfield{author}{\bibinfo{person}{Karen Simonyan} {and} \bibinfo{person}{Andrew Zisserman}.} \bibinfo{year}{2014}\natexlab{}.
\newblock \showarticletitle{Very deep convolutional networks for large-scale image recognition}.
\newblock \bibinfo{journal}{\emph{arXiv preprint arXiv:1409.1556}} (\bibinfo{year}{2014}).
\newblock


\bibitem[Sohn(2016)]%
        {sohn2016improved}
\bibfield{author}{\bibinfo{person}{Kihyuk Sohn}.} \bibinfo{year}{2016}\natexlab{}.
\newblock \showarticletitle{Improved deep metric learning with multi-class n-pair loss objective}.
\newblock \bibinfo{journal}{\emph{Advances in neural information processing systems}}  \bibinfo{volume}{29} (\bibinfo{year}{2016}).
\newblock


\bibitem[Song et~al\mbox{.}(2018)]%
        {song2018rapid}
\bibfield{author}{\bibinfo{person}{Wei Song}, \bibinfo{person}{Yan Wang}, \bibinfo{person}{Dongmei Huang}, {and} \bibinfo{person}{Dian Tjondronegoro}.} \bibinfo{year}{2018}\natexlab{}.
\newblock \showarticletitle{A rapid scene depth estimation model based on underwater light attenuation prior for underwater image restoration}. In \bibinfo{booktitle}{\emph{Advances in Multimedia Information Processing--PCM 2018: 19th Pacific-Rim Conference on Multimedia, Hefei, China, September 21-22, 2018, Proceedings, Part I 19}}. Springer, \bibinfo{pages}{678--688}.
\newblock


\bibitem[Song et~al\mbox{.}(2020)]%
        {song2020score}
\bibfield{author}{\bibinfo{person}{Yang Song}, \bibinfo{person}{Jascha Sohl-Dickstein}, \bibinfo{person}{Diederik~P Kingma}, \bibinfo{person}{Abhishek Kumar}, \bibinfo{person}{Stefano Ermon}, {and} \bibinfo{person}{Ben Poole}.} \bibinfo{year}{2020}\natexlab{}.
\newblock \showarticletitle{Score-based generative modeling through stochastic differential equations}.
\newblock \bibinfo{journal}{\emph{arXiv preprint arXiv:2011.13456}} (\bibinfo{year}{2020}).
\newblock


\bibitem[Szegedy et~al\mbox{.}(2015)]%
        {szegedy2015going}
\bibfield{author}{\bibinfo{person}{Christian Szegedy}, \bibinfo{person}{Wei Liu}, \bibinfo{person}{Yangqing Jia}, \bibinfo{person}{Pierre Sermanet}, \bibinfo{person}{Scott Reed}, \bibinfo{person}{Dragomir Anguelov}, \bibinfo{person}{Dumitru Erhan}, \bibinfo{person}{Vincent Vanhoucke}, {and} \bibinfo{person}{Andrew Rabinovich}.} \bibinfo{year}{2015}\natexlab{}.
\newblock \showarticletitle{Going deeper with convolutions}. In \bibinfo{booktitle}{\emph{Proceedings of the IEEE conference on computer vision and pattern recognition}}. \bibinfo{pages}{1--9}.
\newblock


\bibitem[Tang et~al\mbox{.}(2023)]%
        {tang2023underwater}
\bibfield{author}{\bibinfo{person}{Yi Tang}, \bibinfo{person}{Hiroshi Kawasaki}, {and} \bibinfo{person}{Takafumi Iwaguchi}.} \bibinfo{year}{2023}\natexlab{}.
\newblock \showarticletitle{Underwater image enhancement by transformer-based diffusion model with non-uniform sampling for skip strategy}. In \bibinfo{booktitle}{\emph{Proceedings of the 31st ACM international conference on multimedia}}. \bibinfo{pages}{5419--5427}.
\newblock


\bibitem[Tittensor et~al\mbox{.}(2010)]%
        {tittensor2010global}
\bibfield{author}{\bibinfo{person}{Derek~P Tittensor}, \bibinfo{person}{Camilo Mora}, \bibinfo{person}{Walter Jetz}, \bibinfo{person}{Heike~K Lotze}, \bibinfo{person}{Daniel Ricard}, \bibinfo{person}{Edward~Vanden Berghe}, {and} \bibinfo{person}{Boris Worm}.} \bibinfo{year}{2010}\natexlab{}.
\newblock \showarticletitle{Global patterns and predictors of marine biodiversity across taxa}.
\newblock \bibinfo{journal}{\emph{Nature}} \bibinfo{volume}{466}, \bibinfo{number}{7310} (\bibinfo{year}{2010}), \bibinfo{pages}{1098--1101}.
\newblock


\bibitem[Wang et~al\mbox{.}(2023a)]%
        {wang2023underwater}
\bibfield{author}{\bibinfo{person}{Ning Wang}, \bibinfo{person}{Tingkai Chen}, \bibinfo{person}{Xiangjun Kong}, \bibinfo{person}{Yanzheng Chen}, \bibinfo{person}{Rongfeng Wang}, \bibinfo{person}{Yongjun Gong}, {and} \bibinfo{person}{Shiji Song}.} \bibinfo{year}{2023}\natexlab{a}.
\newblock \showarticletitle{Underwater attentional generative adversarial networks for image enhancement}.
\newblock \bibinfo{journal}{\emph{IEEE Transactions on Human-Machine Systems}} \bibinfo{volume}{53}, \bibinfo{number}{3} (\bibinfo{year}{2023}), \bibinfo{pages}{490--500}.
\newblock


\bibitem[Wang et~al\mbox{.}(2023b)]%
        {wang2023stylediffusion}
\bibfield{author}{\bibinfo{person}{Zhizhong Wang}, \bibinfo{person}{Lei Zhao}, {and} \bibinfo{person}{Wei Xing}.} \bibinfo{year}{2023}\natexlab{b}.
\newblock \showarticletitle{Stylediffusion: Controllable disentangled style transfer via diffusion models}. In \bibinfo{booktitle}{\emph{Proceedings of the IEEE/CVF International Conference on Computer Vision}}. \bibinfo{pages}{7677--7689}.
\newblock


\bibitem[Wu and He(2018)]%
        {wu2018group}
\bibfield{author}{\bibinfo{person}{Yuxin Wu} {and} \bibinfo{person}{Kaiming He}.} \bibinfo{year}{2018}\natexlab{}.
\newblock \showarticletitle{Group normalization}. In \bibinfo{booktitle}{\emph{Proceedings of the European conference on computer vision (ECCV)}}. \bibinfo{pages}{3--19}.
\newblock


\bibitem[Yang and Sowmya(2015)]%
        {yang2015underwater}
\bibfield{author}{\bibinfo{person}{Miao Yang} {and} \bibinfo{person}{Arcot Sowmya}.} \bibinfo{year}{2015}\natexlab{}.
\newblock \showarticletitle{An underwater color image quality evaluation metric}.
\newblock \bibinfo{journal}{\emph{IEEE Transactions on Image Processing}} \bibinfo{volume}{24}, \bibinfo{number}{12} (\bibinfo{year}{2015}), \bibinfo{pages}{6062--6071}.
\newblock


\bibitem[Zhang et~al\mbox{.}(2024)]%
        {zhang2024synergistic}
\bibfield{author}{\bibinfo{person}{Dehuan Zhang}, \bibinfo{person}{Jingchun Zhou}, \bibinfo{person}{Chunle Guo}, \bibinfo{person}{Weishi Zhang}, {and} \bibinfo{person}{Chongyi Li}.} \bibinfo{year}{2024}\natexlab{}.
\newblock \showarticletitle{Synergistic multiscale detail refinement via intrinsic supervision for underwater image enhancement}. In \bibinfo{booktitle}{\emph{Proceedings of the AAAI conference on artificial intelligence}}, Vol.~\bibinfo{volume}{38}. \bibinfo{pages}{7033--7041}.
\newblock


\bibitem[Zhang et~al\mbox{.}(2021)]%
        {zhang2021underwater}
\bibfield{author}{\bibinfo{person}{Tingting Zhang}, \bibinfo{person}{Yujie Li}, {and} \bibinfo{person}{Shinya Takahashi}.} \bibinfo{year}{2021}\natexlab{}.
\newblock \showarticletitle{Underwater image enhancement using improved generative adversarial network}.
\newblock \bibinfo{journal}{\emph{Concurrency and Computation: Practice and Experience}} \bibinfo{volume}{33}, \bibinfo{number}{22} (\bibinfo{year}{2021}), \bibinfo{pages}{e5841}.
\newblock


\bibitem[Zhang et~al\mbox{.}(2018)]%
        {zhang2018shufflenet}
\bibfield{author}{\bibinfo{person}{Xiangyu Zhang}, \bibinfo{person}{Xinyu Zhou}, \bibinfo{person}{Mengxiao Lin}, {and} \bibinfo{person}{Jian Sun}.} \bibinfo{year}{2018}\natexlab{}.
\newblock \showarticletitle{Shufflenet: An extremely efficient convolutional neural network for mobile devices}. In \bibinfo{booktitle}{\emph{Proceedings of the IEEE conference on computer vision and pattern recognition}}. \bibinfo{pages}{6848--6856}.
\newblock


\bibitem[Zhao et~al\mbox{.}(2016)]%
        {zhao2016loss}
\bibfield{author}{\bibinfo{person}{Hang Zhao}, \bibinfo{person}{Orazio Gallo}, \bibinfo{person}{Iuri Frosio}, {and} \bibinfo{person}{Jan Kautz}.} \bibinfo{year}{2016}\natexlab{}.
\newblock \showarticletitle{Loss functions for image restoration with neural networks}.
\newblock \bibinfo{journal}{\emph{IEEE Transactions on computational imaging}} \bibinfo{volume}{3}, \bibinfo{number}{1} (\bibinfo{year}{2016}), \bibinfo{pages}{47--57}.
\newblock


\end{thebibliography}

\end{document}